\documentclass[]{interact}

\usepackage{epstopdf}
\usepackage[caption=false]{subfig}
\usepackage[hyphens]{url}
\usepackage{hyperref}
\usepackage[T1]{fontenc} 

\usepackage{orcidlink}

\usepackage[natbibapa,nodoi]{apacite}
\theoremstyle{plain}

\theoremstyle{definition}

\theoremstyle{remark}

\begin{document}

\articletype{PREPRINT}

\title{Detecting undisclosed LLM-generated content in parliamentary texts}

\author{
\name{Minerva Suvanto\orcidlink{0009-0003-1751-151X}\textsuperscript{a}\thanks{CONTACT Minerva Suvanto. Email: minerva.suvanto@chalmers.se}, 
Andrea McGlinchey\orcidlink{0009-0001-8436-237X}\textsuperscript{b}, 
Peter J\@. Barclay\orcidlink{0009-0002-7369-232X}\textsuperscript{b}, 
Mattias Wahde\orcidlink{0000-0001-6679-637X}\textsuperscript{a}}
\affil{\textsuperscript{a}Chalmers University of Technology, Gothenburg, Sweden \\
\textsuperscript{b}Edinburgh Napier University, Edinburgh, UK}
}

\maketitle

\begin{abstract}
In this paper, we evaluate the extent of undisclosed LLM-generated content in texts from the parliaments of the United Kingdom and Sweden. In many areas, such as in journalism or in academic writing, there are often requirements to clearly disclose whether AI tools, such as LLMs, have been used. In the case of parliamentary texts, the guidelines on disclosure of AI use are more vague. However, in order to maintain transparency and retain public trust, it is generally recommended that parliamentarians should state whether or not they have used AI when writing texts, such as parliamentary motions. Here, we train an interpretable (glass-box) text classifier using pre-LLM parliamentary texts and LLM-generated versions of such texts. We then apply the classifier to a test set containing recent parliamentary texts, finding a steady increase in undisclosed LLM use, in both parliaments, from 2022 onwards.
\end{abstract}


\section{Introduction and motivation}
\label{sect:introduction}

The advent of large language models (LLMs), and the associated chatbots such as ChatGPT, Gemini, Llama, 
and so on, has completely transformed human-computer interaction (HCI), making it easy to generate convincing and coherent human-like text based on instructions (prompts) given in natural language. While there are many excellent uses of LLMs, such as assisting people with writing tasks on almost any topic, there are also many potential risks and drawbacks with their use. Some concerns include the efficient generation and distribution of erroneous texts either accidentally (misinformation) or deliberately (disinformation), or propagation of texts exhibiting discrimination and biases against individuals or groups, \textit{e.g}., minorities. Whether generated with malicious intent or unintentionally, such texts may have harmful consequences both on an individual level and societally.

In many cases, such as in journalism and academic writing, it is often required that authors should clearly state whether artificial intelligence (AI),
typically in the form of LLMs, was used in the process. For example, the guidelines for some mass media organizations and academic journals state that AI use should be disclosed, including a description of how AI was used and which specific models were employed.
In the field of politics, which is the topic of this paper, there are also strong recommendations to disclose the use of AI, further discussed in Section~\ref{sect:politics} below. 

However, such disclosure is largely missing from officially published parliamentary materials, making it difficult to assess to what degree politicians, and their speech writers and other aides, make use of AI when writing.
The lack of disclosure of AI use in parliamentary texts, whether accidental or deliberate, may have a negative societal impact, by (further) eroding the public's trust in politicians, if and when undisclosed AI use comes to light.
Hence, in this paper, we study the frequency of LLM-generated texts in parliamentary statements, both from the United Kingdom (the parliament at Westminster) and from Sweden (the Riksdag), as detected by a glass-box text classifier that has been used also in other studies~\citep{WahdeEtal2024,SuvantoEtal2026,SuvantoWahde2026}. 
Use of LLMs is a new form of HCI that is already ubiquitous and increasingly uncontroversial in many,
if not most, contexts; our aim is not to criticise the use of LLMs~\textit{per se}, but
rather to identify and highlight the \textit{undisclosed} use of LLMs in parliaments, 
by applying a fully interpretable (glass-box) text classifier, rather than a
black-box (neural) classifier.
The paper is structured as follows: In 
Section~\ref{sect:politics} we discuss various
ethical concerns in regards to AI use in politics. Next, in Section~\ref{sect:relatedwork} we provide a description of related work. We then describe, in Section~\ref{sec:data}, the data collection procedure, as well as the method for obtaining LLM-generated data. Next, the classification method is described, in Section~\ref{sect:method}. Results are given in Section~\ref{sect:results}, followed by a discussion Section~\ref{sect:discussion}, and concluding remarks in Section~\ref{sect:conclusion}.

\section{AI in political texts}
\label{sect:politics}

The use of AI in a political context has raised concerns among voters~\citep{tim_harford_bbc_2025} and politicians have already been criticised for using LLM-based tools~\citep{Min2025}, such as ChatGPT, in their work, without always disclosing that they have done so.
Public opinion towards the use of AI in parliamentary decision-making was studied by~\cite{PickeringEtal2025}, who found that nearly half of UK respondents were not supportive of parliamentarians using AI assistants. In Sweden, the report from the~\cite{SOM206} measured public opinions on AI, demonstrating that respondents are concerned about AI-generated misinformation influencing democratic elections. It is  thus clear that the use of AI by political representatives is an important matter for voters. Nevertheless, so far, politicians rarely disclose using AI-tools. 

The risks related to LLMs in politics include, for example, the potential spread of misinformation or discrimination.
In a survey from the~\cite{EBU2025}, over 3,000 AI responses were evaluated by professional journalists. Nearly half of the responses had at least one significant problem
and a fifth of the responses contained major errors regarding the accuracy of information.
\cite{AdamsEtal2023} summarise the effects of misinformation in several disciplines, including politics. They conclude that distributing misinformation is perceived as a threat to democracy, owing to the potential negative effects such as, for example, polarization or making misinformed decisions based on inaccurate information. Misinformation can, of course, be produced by politicians without the use of AI, but as LLMs are known to often fabricate facts and are capable of generating politically persuasive messages~\citep{BaiEtal2025}, there is an increased risk of unintentionally spreading misinformation through careless use of LLMs and,
evidently, also a heightened risk of malicious actors spreading disinformation.

Outputs from LLMs have also been shown to contain subtle (and not so subtle) social biases targeting race, ethnicity, gender, or age when creating text~\citep{TaubenfeldEtal2024,WanEtal2023} or images \citep{mansour_sami_case_2023}, and when translating between languages \citep{barclay_investigating_2024}.
Amplifying such biases can lead to unfair treatment of certain groups of people. Additionally, LLMs can be politically biased, as shown by~\cite{RettenbergerEtal2025}, who studied political bias in the case of open-source language models. As a result of AI use, political bias could be introduced into generated text in ways that contradict the values of the political party that a (hypothetical) politician represents.

Finally, there are, in fact, some guidelines regarding the use of AI in politics.
For example, the Interparliamentary Union (IPU) states in its \textit{Guidelines for AI in parliaments}~\citep{IPU1} that parliaments \textit{\lq\lq must prioritize transparency and responsibility \ldots\rq\rq}~and therefore \textit{\lq\lq label AI-assisted documents, specifying the tool and version used\rq\rq}. Moreover, in 
\textit{Guidelines for AI in parliaments - Key points for MPs}~\citep{IPU2}, the IPU lists transparency, meaning \textit{\lq\lq clear communication about when and how AI is being used\rq\rq}~as an essential 
principle. Furthermore, \textit{\lq\lq supporting transparent communication about AI use\rq\rq}~is mentioned as
a key responsibility for MPs. ‌In the case of the UK, there is the \textit{Generative AI framework for the UK government}, which states, as a recommendation to (all) government employees, to \textit{\lq\lq clearly signpost when generative AI has been used to create content \ldots\rq\rq}~and \textit{\lq\lq where possible, label AI generated content \ldots\rq\rq}~\cite{UKParliament1}.
While such guidelines are typically in the form of recommendations and advice rather than mandatory requirements, their existence nevertheless highlights the relevance of disclosing AI use in parliamentary texts. 

\section{Related work}
\label{sect:relatedwork}
Encountering LLM-generated texts has become increasingly common, but recognising them as such is difficult for humans evaluators, who typically identify LLM-generated text essentially at chance levels (with an accuracy of 0.50 - 0.55)~\citep{SuvantoEtal2026,McGlinchey2025,Clark2021}. The results in~\citep{Richter2026} indicate that human evaluators are slightly better at accurately identifying human-written texts (with an accuracy of 0.67) rather than LLM-generated texts (with an accuracy as low as 0.32). However, there are often differences between human-authored and LLM-generated texts, both in lexical choices and in grammar, which automated LLM-detection methods can leverage~\citep{SuvantoEtal2026,ZamaraevaEtal2025,ReinhartEtal2025,Munoz2024,Zeleke2025,Herbold2023}.

A survey from~\cite{WuEtal2025} introduces methods used for the detection of LLM-generated text, categorising current methods into (i) watermarking technologies, (ii) statistics-based methods, and (iii) neural-based methods. Watermarking technologies rely on injecting identifiable patterns in the model output, for example, by promoting the use of certain tokens. A text can be watermarked by the model during generation or it can be done after the fact by post-processing the output. The watermark can then be detected using various detection algorithms. This method relies on the deployer of the LLM to provide the watermarking technology. Statistics-based and neural-based methods, on the other hand, use patterns extracted directly from text in order to detect LLM use. Neural-based methods typically use deep neural networks, such as pre-trained LLMs, to perform end-to-end detection of texts.

Unwarranted accusations of inappropriate AI use can potentially damage the reputation of an accused individual.
For example, false accusations of AI plagiarism can be harmful to students' or scholars' academic careers~\citep{Giray2024, Dalalah2023} and cause significant psychological stress~\citep{Gegg2024}. ~\cite{Giray2024} 
and~~\cite{Gegg2024} also suggest that AI-detection tools are biased against marginalised groups, such as second-language learners or neurodivergent writers.
In such high-stakes scenarios, AI detection tools should be used with caution, and it is crucial to understand the underlying \textit{reasons} for a positive detection.
The use of interpretable (glass-box) models is essential in such situations~\citep{Rudin2019}, as their decision-making is transparent. \cite{SuvantoEtal2026} employ an explicitly interpretable, linear classifier in the domain of creative writing. Their analysis reveals some prominent features for the detection task but, on the other hand, shows that a decision is typically based on a large constellation of features, making it challenging to evade a positive detection
by making a few targeted modifications in an otherwise LLM-generated text.
A method proposed by \cite{QuidwaiEtal2023} measures the similarities between (known) LLM-generated texts and student-written exam answers. In their method, the cosine similarity of sentence pairs is used for measuring the average similarity of a full document. Their method, while simple, obtains a high accuracy, challenging the indiscriminate use of black-box, end-to-end detectors.
~\cite{WeiEtal2025} explore a method based on findings that show how LLMs tend to rephrase human-authored text more heavily compared to LLM-generated texts. In their method, the edit distance between a candidate text and an LLM-rephrased version is used as a basis for the detection, where a larger edit distance suggests that the text was human-written.

Detector performance varies immensely depending on the method used, with some methods reporting near perfect accuracies while others misclassify texts consistently. Metrics reported for different methods are summarised by~\cite{WuEtal2025}. The performance of a detector, and consequently its reliability, is also impacted by factors besides the choice of method. ~\cite{DaSilvaGameiro2024} highlights the poor performance of general detectors designed to work in a broad range of contexts, while detectors for one specific domain can have near perfect reported accuracies. Another aspect to consider for reliable detection is the length of the test samples. As shown in~\cite{VermaEtal2024}
and~~\cite{WuEtal2024}, the performance of the classifier increases with the length of the (test) texts, and shorter texts are more likely to be misclassified. On the other hand, ~\cite{WuEtal2024} show that shorter texts in the \textit{training} set yield better performing detectors. Finally, detector performance can be negatively impacted by adversarial attacks~\citep{ChengEtal2026,WuEtal2025} where the output text from an LLM is modified so as to avoid detection.

Following a common trend in NLP research, much focus is given to detectors trained in English. A few studies do consider specific other languages~\citep{Salem2026,Mao2025} or multilingual benchmarking data sets~\citep{Macko2023,Wang2024}. However, in Swedish, one of the languages considered in this paper, only some commercial tools for detecting LLM-generated text exist and, to the best of our knowledge, there are no publicly available LLM detectors or data sets for training detectors exclusively in Swedish.

For the specific case considered here, LLM usage in a parliamentary texts, various aspects have been studied.
Some effort has been made to study the style and content of political texts produced by LLMs, for example, their tendency to elicit specific political leanings~\cite{Rozado2024} as well as their political persuasiveness~\citep{PotterEtal2024,HackenburgEtal2025}. \cite{VonDerHeyde2025} also noted that biases in LLMs make them unsuitable for predicting voter behaviour. Evaluating the quality of parliamentary text generated by LLMs has been studied by~\cite{KoniarisEtAl2025}, who proposed various evaluation benchmarks which were then applied to evaluate whether the quality of LLM-generated parliamentary texts can be improved via fine-tuning.

The specific task of \textit{detecting} LLM-generated parliamentary text, however, is less studied. To the best of our knowledge, the only other study to consider this topic is from~\cite{SuvantoWahde2026}. In their work, the performance of an LLM detector is only evaluated on a labelled test set of political speeches (extracted from the data set introduced by~\cite{KoniarisEtAl2025}). 

\section{Data sets}
\label{sec:data}
Parliamentary texts come in different varieties, either as written text, such as
prepared statements (\textit{e.g}., parliamentary motions), and as spoken language in
the form of transcribed speech.
In both cases, the texts can be fully
or partially LLM-generated (for instance, speakers may recite their speech from a manuscript generated by an LLM). While transcribed texts are typically cleaned by removing filler words, correcting grammar, and so on, they may still contain traces of spontaneous speech. 
While LLMs are generally capable of generating human-like speech, deliberately prompting them to replicate spontaneous speech cues can be challenging.
Thus, we have chosen to use texts in written language instead of transcribed speeches in order to generate a high-quality data set for training a classifier for the detection task.

In this work, we consider texts in two languages, namely, British English (UK) and Swedish (SWE). The human-written texts for these languages were obtained from the UK Parliament and the Riksdag (Sweden's parliament), as described in Subsections~\ref{subsec:dataUK} and \ref{subsec:dataSWE} respectively. The LLM-generated texts were obtained by prompting two models (Gemini and GPT) for full texts, based on summaries. The procedure is described in Subsection~\ref{subsec:dataGeneration}.

The construction and preprocessing of the data sets (one with data from the UK, and one with Swedish data) used for training and evaluating the classifier (referred to hereafter as the \textit{base data sets}) is described in Subsection~\ref{subsec:standardData}. 
We consider parliamentary texts from the years 2014 - 2020 as (fully) human-written, since they were published prior to the era of modern LLMs\footnote{While LLMs did, in fact, exist before the release of ChatGPT in Nov.~2022, they were mainly employed by specialists and using them required some technical expertise. Therefore, it is highly unlikely that parliamentary texts published before 2022 would be LLM-generated.}. 
Texts published in 2021 and onward (for cut-off dates, see the following subsections) are instead used for the task of investigating the real-world use of LLMs by parliamentarians. These data sets (again, one for the UK and one for Sweden) are henceforth referred to as the \textit{forward test data sets}, described in Subsection~\ref{subsec:forwardData}. 
In these sets, the texts from 2021 and most of 2022 predate the public release of LLMs, and are held-out for measuring false positive (FP) rates. Texts from late 2022 and onwards do not have ground-truth labels: Some may be LLM-generated (in part or in whole) whereas other may be fully human-written, and it is the task of our classifier, trained over the respective base sets, to detect LLM usage in this time period, as described below.
\subsection{UK data}
\label{subsec:dataUK}

We extracted publicly available parliamentary statements\footnote{Source: Hansard, \url{https://hansard.parliament.uk/}. Contains Parliamentary information licensed under the Open Parliament Licence v3.0. License: \url{https://www.parliament.uk/site-information/copyright/open-parliament-licence/}} written in standard British English. We extracted data for the base set from the beginning of 2014 until the end of 2020, and for the forward set from the beginning of 2021 until 29 April 2026. The following column headings were extracted for each written statement: "URL", "Heading", "Date", "Speaker Title", "Speaker Name" and "Statement". The column location information added by the column indicator widget was cleaned from the text, and additional parentheses were stripped as needed. All HTML tags were removed, so that only the plain text was stored. The Date field had to be extracted from a larger string, for example "Volume 634: debated on Monday 8 January 2018". Paragraph breaks were retained using "\verb+\n\n+" as separators between paragraphs; however, all other spacing formats were merged to single spaces and the leading and trailing spaces were removed. In total we extracted 9,475 files; the base set contained 5,266 files and the forward set had 4,209 files. 

\subsection{Swedish data}
\label{subsec:dataSWE}
We manually downloaded publicly available parliamentary motions, \textit{i.e.}, suggestions from members of parliament, in text format\footnote{Source: The Swedish Parliament,  see \url{www.riksdagen.se/sv/dokument-och-lagar/riksdagens-oppna-data/dokument/} and \url{https://www.riksdagen.se/sv/dokument-och-lagar/riksdagens-oppna-data/anvandarstod/anvandningsvillkor/}} from the Swedish Riksdag in accordance with the stated usage policies. Data were downloaded 
in chunks covering the parliamentary sessions for 2014-2015, 2015-2016, \ldots, 2025-2026. The base set (see above) thus involves data up to and including mid-2021, \textit{i.e.}, the end of the 2020-2021 parliamentary session. The forward test set contains data from the autumn of 2021 until 13 April 2026 (the cut-off date for the download). In total, the raw data for the base set contained 30,077 files. These files were then preprocessed individually, removing headers and signatures. Then, both short files (less than 2,000 characters) and
very long files (more than 5,000 characters) were removed as well, resulting in a total of 7,778 files. The forward test set, initially with 28,519 files, was preprocessed in the same way except that, here, the length filtering only removed short texts (again with a cut-off at 2,000 characters), resulting in a total of 13,565 files.

\subsection{Text generation procedure}
\label{subsec:dataGeneration}
Generating texts via LLMs can be done, for example, by prompting a model to rewrite snippets of texts. Here, however, we use a different approach, bearing in mind how a parliamentarian may realistically use LLMs to produce texts, typically starting with only a short summary of the key points of a desired text.

Zero-shot prompting was used for obtaining LLM-generated texts based on summaries of human-written texts, as follows: First, we prompted an LLM to generate summaries from full human-written texts (of the type described above), retaining the key message of the original text. Then, for each summary, we prompted an LLM (in a new session) to generate a full text from the summary. 

Specifically, we selected random subsets of 200 texts (that were at least 150 tokens long) from each time period to summarise, from both the UK and SWE data sets, resulting in 1,400 texts for each language. For the summarisation, we used the model Gemini 3.1 Flash-Lite via Gemini's API with its default model parameters. The summaries were sampled (manually) and were found to be of high quality, both grammatically and semantically. We also experimented with the models Llama 3.1 and Gemma 3, but found that the quality of the generated texts was rather poor, particularly in Swedish. 

Next, in order to generate the full texts, we used the same Gemini model for the UK and SWE data sets to produce 1,400 LLM-generated texts 
(one for each summary). For the UK set, we also used GPT-5 mini via OpenAI's API to obtain an additional 1,400 LLM-generated texts.
We tested various prompts for both summarisation and full generation and sampled and manually evaluated the outputs. We selected the prompts that produced the best quality outputs for our experiments; the final prompts used for generating the summaries and the full texts are included in Appendix~\ref{appendix:prompts}. 

\subsection{Base data set}
\label{subsec:standardData}

A base data set used for training and evaluating a binary text classifier was constructed from both the human-written texts (Class~0) and LLM-generated texts (Class~1). As mentioned above, during the text generation procedure (described in the previous subsection), 1,400 human-written texts per language were summarised. Thus, to ensure that our data set did not have a significant class imbalance, we included these 1,400 human-written texts and the same number of LLM-generated texts obtained from their summaries in the base data set. For the Swedish texts, all the LLM-generated texts were obtained from 
Gemini. For the UK texts, we included 700 from Gemini and 700 from GPT.

The texts were preprocessed and tokenised using an inclusive tokeniser that splits words on punctuation marks and whitespaces, maintaining word forms and casings. We also replaced URLs with a placeholder token. Punctuation marks were completely removed so that they would
not influence the detection; as also noted in~\cite{SuvantoWahde2026}, there is a small positive effect in detection when including such features in the text. An overview of the data sets after preprocessing is shown in the Table~\ref{tab:standardData}.

The preprocessed texts were then divided into paragraphs, using newline characters as a paragraph separator. The texts may also contain short lines of texts, such as lists of items spanning multiple lines; we only retained paragraphs that have a length $L\geq k$ where $k$ is a minimum length threshold to avoid identifying such items as separate paragraphs. 

\begin{table}
\caption{Text statistics for the subsets in the base data sets. Each subset consists of 1,400 items. For each subset, we measured the average number of characters $\overline{C}$, the average number of tokens $\overline{T}$, the average number of paragraphs $\overline{P}$ and paragraph lengths (as number of tokens) $\overline{P}_{\rm{len}}$, computed after preprocessing. }
\begin{center}
\begin{tabular}{l|c|c|c|c}
    Subset  & $\overline{C}$ & $\overline{T}$    &   $\overline{P}$ &   $\overline{P}_{\rm{len}}$  \\
    \hline
    \multicolumn{3}{l}{UK} \\
    \hline
    Human  (Class 0)  & 3,043.4  & 498.3   & 11.5 & 43.2  \\
    Gemini (Class 1)   &  1,933.4   &  293.7  & 3.3 & 87.7 \\
    GPT  (Class 1)  & 2,316.7  & 353.4 & 6.6 & 53.3 \\
    \hline
    \multicolumn{3}{l}{SWE} \\
    \hline
    Human (Class 0)  &  2,705.3   &  408.6   & 7.8  & 52.2 \\
    Gemini (Class 1)  &  2,262.0   & 315.2   & 4.8  & 65.3 \\
    \hline
\end{tabular}
\label{tab:standardData}
\end{center}
\end{table}

\subsection{Forward test data set}
\label{subsec:forwardData}

The forward test data sets contain more recent texts from 2021 onwards (the exact date cut-offs for each language are described above). 
The purpose of these sets is to estimate the rate of LLM use in parliamentary texts.
The data sets (UK and SWE) were prepared for the classifier by preprocessing and tokenising them exactly as described in Section~\ref{subsec:standardData}. However, the texts were not divided into paragraphs, so that they can be classified either in full (as a single unit), or by considering smaller chunks (for example, paragraphs). In the latter case, a text can be considered LLM-generated if, for example, any of its paragraphs is detected as such.
The size of each forward test set and text statistics (of the preprocessed text) are shown in Table~\ref{tab:forwardData}.

\begin{table}
    \caption{The number of texts $N$, the average number of characters $\overline{C}$, the average number of tokens $\overline{T}$, and the average number of paragraphs $\overline{P}$ in the preprocessed forward test sets.}
    \centering
    \begin{tabular}{l|c|c|c|c|c}
         Language & $N$  & $\overline{C}$ & $\overline{T}$    &   $\overline{P}$&   $\overline{P}_{\rm{len}}$ \\
        \hline
         UK & 4,209 & 3,235.0 & 527.5 & 11.7 & 45.2 \\
         SWE & 13,565 & 4,445.9 & 662.3 & 13.7 & 48.3 \\
        \hline
    \end{tabular}
    \label{tab:forwardData}
\end{table}

\section{Method}
\label{sect:method}

In this study, we have used the linear, explicitly interpretable classification method 
from~\cite{WahdeEtal2024}, to which has been added specific tools for visualizing
and interpreting the reasons for a particular classification decision, as well as
a tool for computing confidence scores, resulting in a method henceforth referred to as  
the \textbf{I}nterpretable \textbf{CON}fidence-enhanced Perceptron (ICON).
The ICON method was recently applied to the detection of LLM-generated creative writing in~\citep{SuvantoEtal2026}. In this approach, the classification of a text is based on the sum of feature weights (referred to as the \textit{classification sum})
\begin{equation}
s = \alpha + \sum_{i=1}^V w_if_i,
\label{eq:s}
\end{equation}
where $\alpha$ is a bias term, $w_i$ are feature weights, and $f_i$ are feature values. A text is labelled as belonging to Class~1 if $s \geq 0$, whereas if $s < 0$, the text is assigned to Class~0.

The features can be of any kind, in principle. Here,
the feature set consists of a set of $n$-grams, with values of $n$ ranging from 1 up to a given
maximum value $n_{\rm{max}}$, extracted from a tokenised training data set. An optional minimum count $c_{\rm{min}}$ is defined to combat overfitting during classifier training. Let $V_1$ denote the number of unigram features in the training set, $V_2$ the number of bigram features, and so on, so that the total number of features becomes
\begin{equation}
    V = \sum_{i=1}^{n_{\rm{max}}}V_i.
\end{equation}
The weight for each feature $f_i$ is initialised in the range $[-1,1]$ as
\begin{equation}
w_i = \frac{c_1(i) - c_0(i)}{c_{1}(i) + c_{0}(i)}
\end{equation}
where $c_0(i)$ and $c_1(i)$ denote the number of occurrences of feature $f_i$ appearing in Class~0 and Class~1, respectively. The bias term is similarly initialised as
\begin{equation}
    \alpha = \frac{N_1 - N_0}{N_1 + N_0} \equiv \frac{N_1 - N_0}{N}.
\end{equation}
where $N_0$ and $N_1$ denote the number of texts in class~0 and class~1, respectively,
and $N$ is the total number of texts in the data set.

The weights are optimised following the training method introduced by~\cite{WahdeEtal2024}. Note that, in this case, we do not use a length-dependent adjustment introduced there. The training method is similar to that of the standard (linear) perceptron, but not identical: In the ICON method, the feature error $e_i$ is computed as

\begin{equation}
e_i = \frac{1}{\gamma_i}\sum_{j=1}^N v_{ij}(\hat{C_j}-C_j), i=1,...,V,
\end{equation}
where $\gamma_i \equiv c_1(i) + c_0(i)$, $v_{ij}$ is the number of occurrences (instances)
of feature $f_i$ in text $j$, whereas $C_j$ and $\hat{C}_{j}$ are the ground
truth and inferred class labels (either 0 or 1, in both cases), respectively.
Thus, unlike the perceptron algorithm where samples are considered one-by-one in
random order, here we compute the error $e_i$ for each feature over the \textit{entire}
training set. The weights are then updated as

\begin{equation}
w_i \leftarrow w_i - \eta e_i,
\end{equation}
where $\eta$ is the learning rate. 
The error for the bias term is defined as
\begin{equation}
e_\alpha = \frac{1}{N}\sum_{j=1}^N(\hat{C_j}-C_j)
\end{equation}
and it is then updated as
\begin{equation}
\alpha \leftarrow \alpha - \eta e_\alpha.
\end{equation}
The training process is iterated for a number of epochs, while also measuring the classifier performance (such as the accuracy or F1 score) over both the training set and the validation set and, in the end, selecting the classifier with the highest validation score. The performance over the validation set is not communicated to the training algorithm in any way, \textit{i.e.}, we follow the standard procedure in holdout validation.

\subsection{Confidence scores}
\label{subsec:confidenceScores}

The absolute value of the classification sum, $|s|$, can be used for measuring the classifier's confidence as described by~\cite{WahdeEtal2024}. 
A confidence histogram is formed by measuring the accuracy over samples in the validation set as a function of the magnitude of the classification sum, \textit{i.e.}, $|s|$. The values of $|s|$ are binned in intervals of width $d$ and the fraction of correctly classified samples (\textit{i.e.}, the accuracy) is computed for the samples falling into each bin. The fraction is then used as the confidence measure, denoted $c(|s|)$, which we use (in Section~\ref{subsec:forwardResults}) for accepting classifications of shorter texts.

\section{Results}
\label{sect:results}
In this section, we present the results obtained with two classifiers, one trained for detecting English texts and one for Swedish texts. In the first Subsection~\ref{subsec:parameterResults}, we cover our procedure for setting the parameters used during training. The evaluation results over the (base) test set are presented in Subsection~\ref{subsec:standardResults}. Finally, in Subsection~\ref{subsec:forwardResults} we present the main results of this paper, involving measures of the rate of LLM-generated texts in the forward test sets.
\subsection{Parameter settings and data splits}
\label{subsec:parameterResults}

As mentioned in Section~\ref{sect:method}, the method involves two parameters, namely the maximum feature ($n$-gram) size $n_{\rm{max}}$ and the minimum inclusion count $c_{\rm{min}}$ for the features. To this set can be added a third parameter, namely a lower limit $k$ (henceforth referred to as the \textit{length threshold}) for including a particular text sample in the data set. Setting a minimum length threshold generally improves the performance of a classifier (over the texts that remain) as shorter texts can be ambiguous and thus be misclassified more often. At the same time, the value of $k$ should not be set too large since a high length threshold will lead to many texts being discarded and may also result in class imbalances due to differences in paragraph counts and lengths between the classes (see Table~\ref{tab:standardData}).

We conducted a set of experiments over the base data set, using different settings of the three parameters. We generated subsets of the base data set with the length threshold $k$ at 10, 20, 30, 40, 50, and 60. We then trained classifiers for each subset with combinations of the parameter values $n_{\rm{max}}=1,2,$ and $3$, and $c_{\rm{min}}=2,10,30,$ and $50$.
 
The criteria for selecting the best value for $k$ were (i) the resulting data set should give the highest validation F1 score, subject to the additional conditions that (ii) the size of the minority class in the data set should be at least 40\% of the total sample size, and (iii) at most 30\% of the samples should be discarded.
The values that satisfied these requirements were $k=30$, $n_{\rm{max}}=2$ and $c_{\rm{min}}=30$ for the UK set and $k=30$, $n_{\rm{max}}=2$ and $c_{\rm{min}}=10$ for the SWE set. The base data sets with the length threshold at $k=30$ consisted of 16,339 samples (71\% of the samples) and 14,413 samples (81\% of the samples) for the UK and SWE sets, respectively. The texts in these two data sets were ordered randomly and then each divided into a training set (80\% of the samples), a validation set (10\%), and a test set (10\%).

\subsection{Classifier performance}
\label{subsec:standardResults}

\begin{figure}[t]
    \centering
    \includegraphics[width=0.49\linewidth]{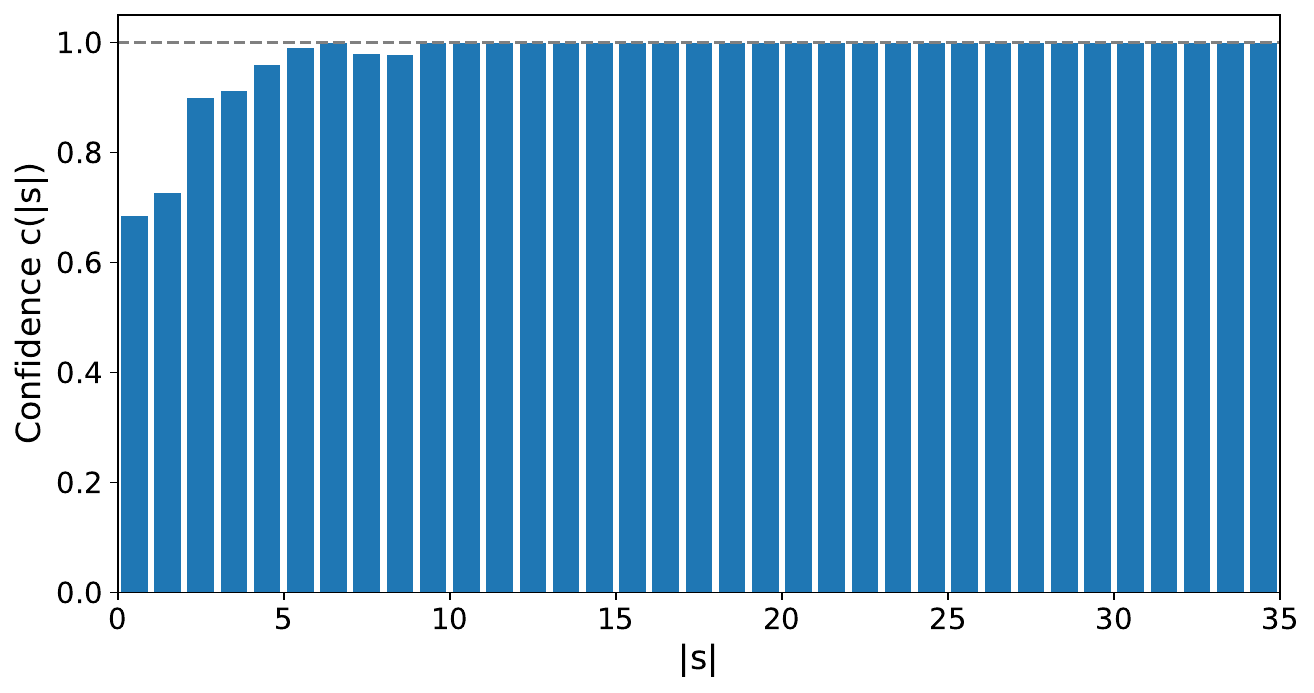}
    \includegraphics[width=0.49\linewidth]{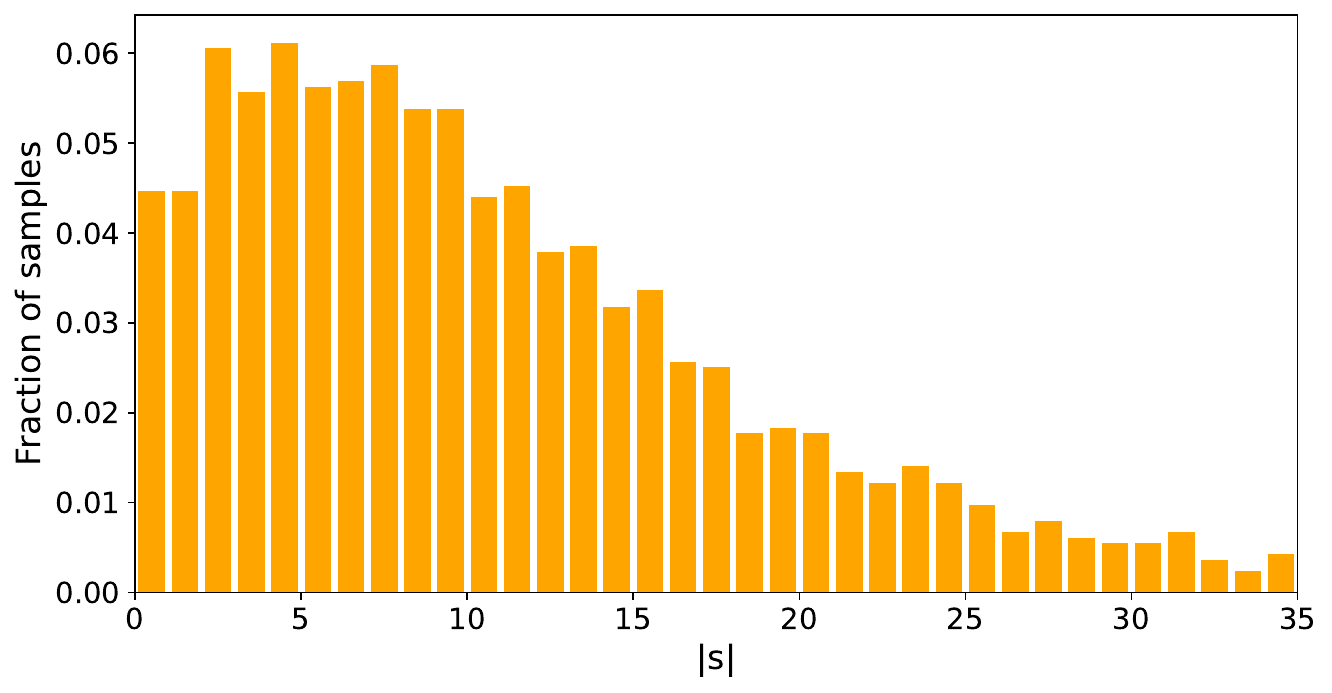}
    \caption{Left Panel: A histogram measuring the confidence score (the fraction of correctly classified text samples) over the validation set from the UK version of the base set, as a function of the (binned) classification sum magnitude, $|s|$; see also Equation~(\ref{eq:s}). Right panel: The fraction of samples in each bin. Note that, for visual clarity, the histograms shown here have been truncated at $|s| = 35$. For bins with $|s| > 35$ (which contain only a small fraction of the total number of texts), the fraction of correctly classified samples is at (or very close to) 1.}
    \label{fig:ukConfidence}
\end{figure}

\begin{figure}
    \centering
    \includegraphics[width=0.49\linewidth]{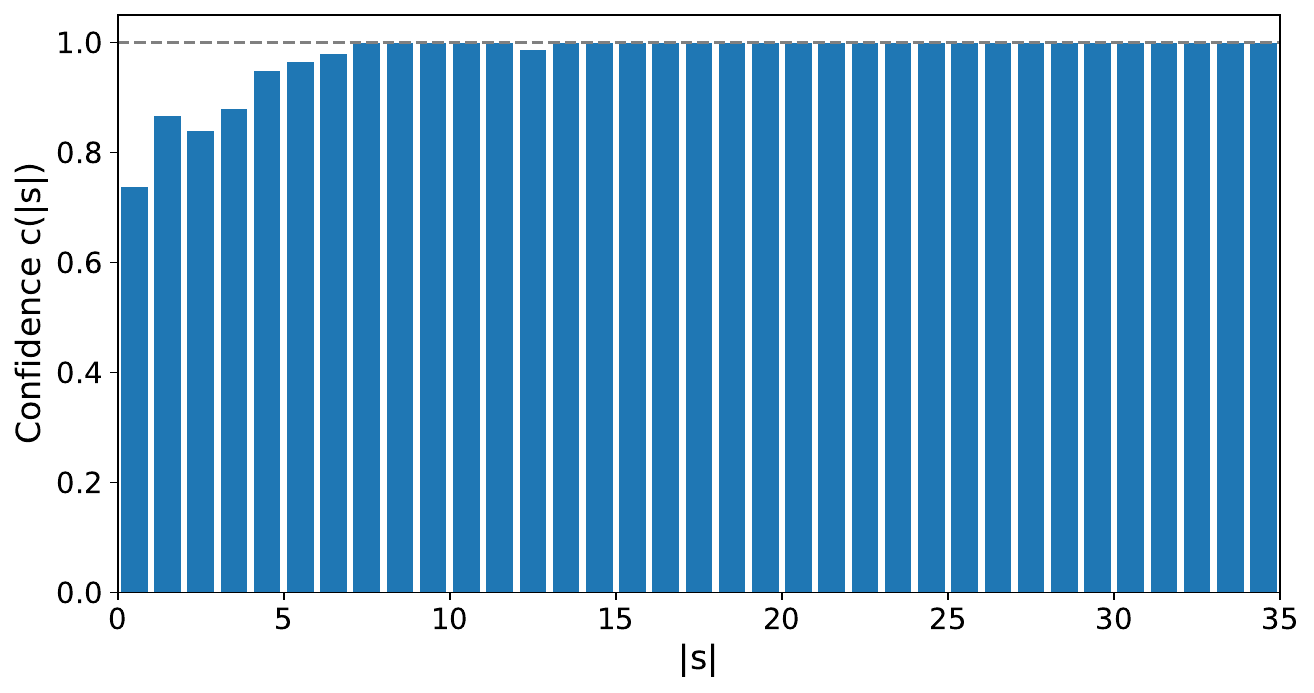}
    \includegraphics[width=0.49\linewidth]{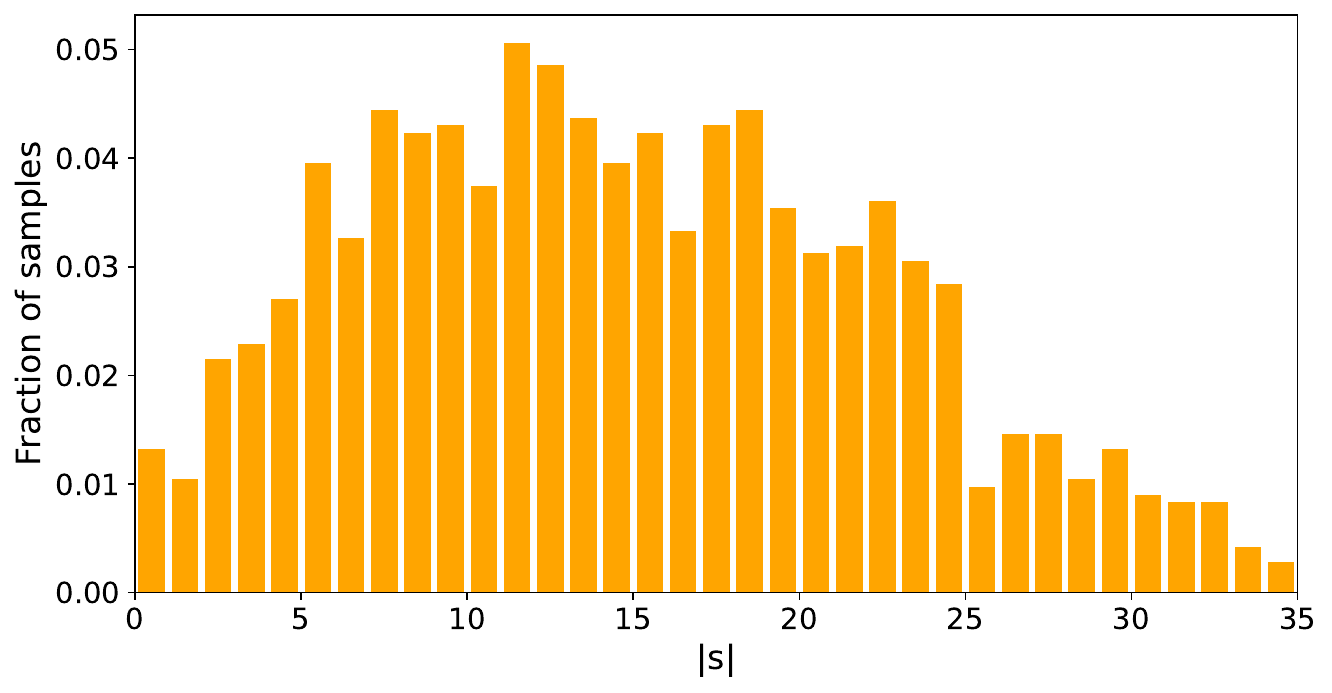}
    \caption{Histogram over the confidence score (left panel) and the fraction of samples (right panel) in each bin, for the Swedish version of the validation set; see also the caption of Figure~\ref{fig:ukConfidence}.}
    \label{fig:sweConfidence}
\end{figure}

Classifiers were trained over the two base sets, \textit{i.e.}, the UK data set and the SWE data set, using the selected parameters for each set (see Subsection~\ref{subsec:parameterResults}).
The classifiers were evaluated over the (base) held out test sets:
The UK classifier obtained an F1 score of 0.940 and an accuracy of 0.950, while the SWE classifier performed slightly better with an F1 score of 0.969 and an accuracy of 0.972. The false positive rates computed over the respective base test sets are 0.062 for the UK data and 0.043 for the SWE data.
Thus, in the vast majority of cases, our trained classifiers can effectively
distinguish human-written paragraphs from LLM-generated ones. Moreover, following the
procedure described in~\cite{WahdeEtal2024}, we generated a confidence measure by
binning the texts in each validation set (UK and SWE, separately) according to their value of the (magnitude of the) classification sum, \textit{i.e.}, $|s|$ (see Equation~(\ref{eq:s}) above). We then computed the fraction of correctly classified samples within each bin, thus obtaining a measure of
classification confidence, shown in Figures~\ref{fig:ukConfidence} and~\ref{fig:sweConfidence}. As can be seen in these figures, for values of $|s|$ beyond 6 or so, the classification accuracy is very close to 1. 


\subsection{Forward test}
\label{subsec:forwardResults}
We considered two approaches for classifying the unlabelled texts. First, we classified the parliamentary motions in full where no paragraphs splits were made and each text was tokenised as one unit. The tokenised texts were then classified as human-written (Class~0) or LLM-generated (Class~1). For this case, one may hope to obtain a low false positive rate (which indeed we find; see below) since the evaluations are based on longer texts than in the case of individual paragraphs.

Second, we investigated whether parliamentary motions contained parts of LLM-generated text by classifying each (tokenised) paragraph individually. Texts where at least one paragraph was labelled as LLM-generated were flagged. Results from both experiments were aggregated by time periods in order to study the trend of using LLMs for parliamentary text over time.

The results of the first experiment are shown in Table~\ref{tab:forwardResultsFull} where one can observe an upward trend in the use of LLMs in both the UK and SWE data sets. The increase for the SWE data is steady over time, whereas in the UK data, there is a sharp increase in the most recent time period. Over the last year (2026 and 2025 - 2026, for the UK and SWE data sets, respectively), the detection rate was 0.021 for the UK data and as high as 0.065 for the Swedish data.

\begin{table}
\caption{Results from classifying parliamentary motions in full, as one unit. The \textit{detection rate} measures the fraction of texts that were classified as LLM-generated in the given time period. The time periods for the UK data are calendar years. For the SWE data, the periods are instead given as working sessions, which begin yearly in autumn (September) and effectively end with a summer break starting in late June.}
\begin{center}
\begin{tabular}{l|r}
    \hline
    \multicolumn{2}{c}{\textbf{UK}} \\
    \hline
    Period  & Detection rate  \\
    \hline
    2021      & 0.0013   \\
    2022      & 0.0013   \\
    2023      & 0.0013   \\
    2024      & 0.0029   \\
    2025      & 0.0023   \\
    2026      & 0.0212   \\
    \hline
\end{tabular}
\quad
\begin{tabular}{l|r}
    \hline
    \multicolumn{2}{c}{\textbf{SWE}} \\
    \hline
    Period  & Detection rate  \\
    \hline
    & \\
    2021 - 2022     &  0.0003  \\
    2022 - 2023     &  0.0011  \\
    2023 - 2024     &  0.0185  \\
    2024 - 2025     &  0.0473  \\
    2025 - 2026     &  0.0647  \\
    \hline
\end{tabular}
\label{tab:forwardResultsFull}
\end{center}
\end{table}

\begin{table}
\caption{Results from classifying parliamentary motions by considering individual paragraphs within each motion.  Here, a motion is considered
LLM-generated if a least one of its paragraphs is detected as such.}
\begin{center}
\begin{tabular}{l|r}
    \hline
    \multicolumn{2}{c}{\textbf{UK}} \\
    \hline
    Period  & Detection rate  \\
    \hline
    2021      & 0.0235 \\
    2022      & 0.0351 \\
    2023      & 0.0444 \\
    2024      & 0.0609 \\
    2025      & 0.0809 \\
    2026      & 0.1545 \\
    \hline
\end{tabular}
\quad
\begin{tabular}{l|r}
    \hline
    \multicolumn{2}{c}{\textbf{SWE}} \\
    \hline
    Period  & Detection rate  \\
    \hline
    & \\
    2021 - 2022     &  0.0099 \\
    2022 - 2023     &  0.0166 \\
    2023 - 2024     &  0.0406 \\
    2024 - 2025     &  0.0799 \\
    2025 - 2026     &  0.0943 \\
    \hline
\end{tabular}
\label{tab:forwardResultsParas}
\end{center}
\end{table}

The first time period for each language (2021 and 2021 - 2022, respectively) predates the public availability of modern LLMs, and can therefore be used to evaluate the false positive rate, which is notably low: Only one full text sample was incorrectly classified as LLM-generated in this time period, for each data set. 

In the second experiment where the classification was based on consideration of individual paragraphs,
two steps were implemented in order to reduce potential false positives.
The minimum length threshold $k=30$ was applied to the paragraphs, similar to the base set (see Subsection~\ref{subsec:parameterResults}). Additionally, a threshold for the classification sum was applied such that paragraphs where $|s|<T$ were not assigned a class label. A suitable threshold $T$ was selected using the confidence histograms in Figures~\ref{fig:ukConfidence} and~\ref{fig:sweConfidence}.
The lowest value of the bin with a confidence measure $c(|s|)\geq0.98$ was selected as the threshold, which was $T=5.0$ for the UK data and $T=7.0$ for the SWE data. 
There were 155 texts (3.7\% of the texts) in the UK data set and 288 texts (2.1\%) in the SWE data set that did not fulfil these requirements and were therefore not classified. 

The results from the second experiment are shown in Table~\ref{tab:forwardResultsParas}. An upward trend in texts classified as (partially) LLM-generated can be seen here as well. 
While the false positive rate here is somewhat higher than for the full (longer) texts, the interesting aspect to consider is the steady increase of LLM-generated texts for both languages. One can also see the same sharp increase for the most recent time period 2026 for the UK data.

We also checked whether any of the texts disclosed LLM use by searching the texts for keywords of popular LLMs and related terms: \textit{ai, chatbot, chat bot, chatgpt, claude, co pilot, copilot, gemini, gpt, grok, language model, llama, llm}. Texts that contained these keywords were manually checked; \textit{none} of them disclosed LLM use.

\section{Discussion}
\label{sect:discussion}

\begin{figure}
    \centering
    \includegraphics[width=0.49\linewidth]{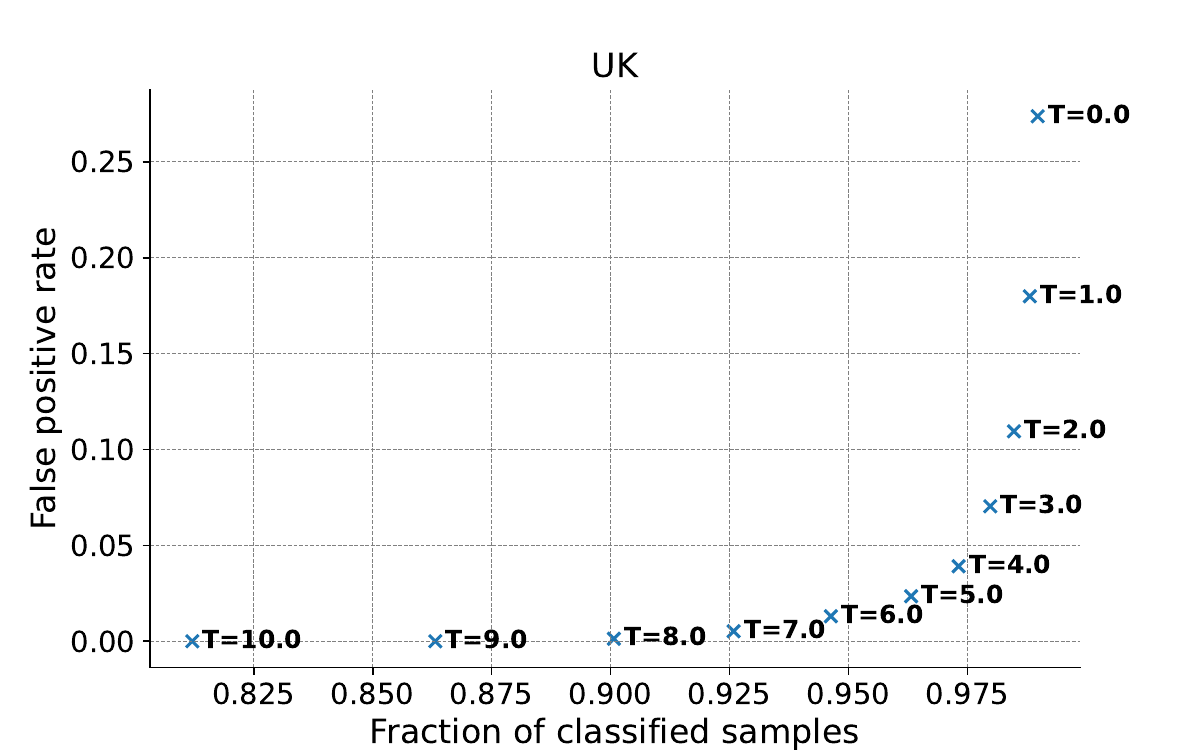}
    \includegraphics[width=0.49\linewidth]{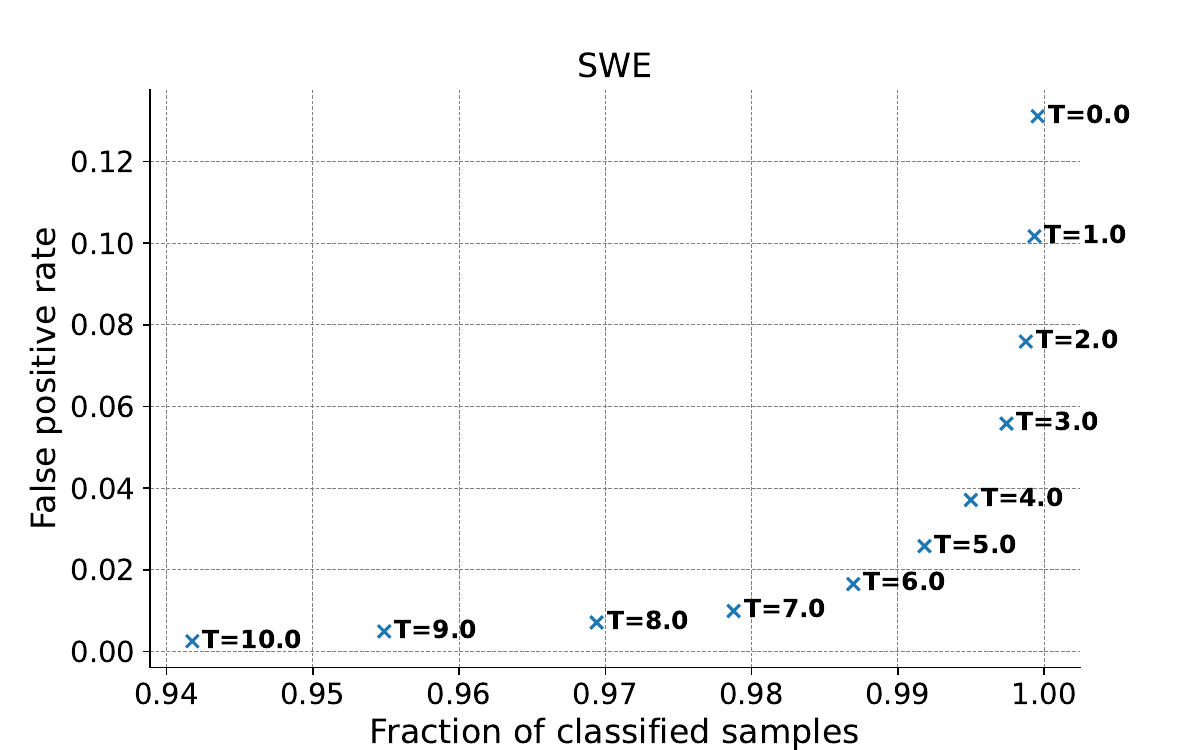}
    \caption{The relation between the false positive rate (measured for the subset of the forward
    test sets covering the year 2021 for the UK and the 2021-2022 parliamentary session for Sweden) 
    and the fraction of classified samples, as the threshold $T$ is varied. The left panel shows the results
    for the UK data and the right panel shows the results for the SWE data.}
    \label{fig:fprates}
\end{figure}

As our main result, we detect a significant, and rapidly growing, fraction of LLM-generated text in parliamentary motions, both on the level of paragraphs within the full texts (Table~\ref{tab:forwardResultsParas}) and, even more interestingly, on the level of full texts considered as one unit (Table~\ref{tab:forwardResultsFull}).
In the latter case, the detection indicates that a significant part of the text is LLM-written (or, at the very least, significantly edited by an LLM), though not necessarily the entire text.
We also observe that \textit{none} of the texts, English or Swedish, contain \textit{any} form of disclosure regarding LLM usage. If our detections are correct, which we have every reason to believe, especially in the case of full texts (Table~\ref{tab:forwardResultsFull}) where our measured false positive rate is close to zero, this non-disclosure is 
an egregious violation of the guidelines (none of which are binding or mandatory, however) established by various organizations that oversee government work. 

As mentioned in Sections~\ref{sect:introduction} and~\ref{sect:method}, we have chosen to use
the straightforward, linear, and fully transparent ICON classifier, rather than a 
neural (black-box) alternative, partly in view of its performance in other studies (see, \textit{e.g.},~\cite{WahdeEtal2024} and~\cite{SuvantoEtal2026}), and partly in keeping with the
idea of maintaining transparency in high-stakes decision-making: If a text, in whatever context, is flagged as being LLM-generated, it is (in our view) essential that one should be able to inspect, visualise, and understand the underlying reasons for the classification, for example to reduce the risk of false accusations. As noted in Section~\ref{sect:results}, the ICON classifier does indeed perform well also in the case studied here, particularly in avoiding false positives, especially in cases involving full-text classification (Table~\ref{tab:forwardResultsFull}). While it is possible
that a neural, black-box classifier could reach a slightly better performance, it would do so at the price of near-complete opaqueness regarding the underlying reasons for its classification decisions.

For the paragraph-level classification (Table~\ref{tab:forwardResultsParas}), it is instructive 
to study how the FP rate varies with the classification
threshold $T$. A higher threshold should give a lower FP rate, but will also reduce the
fraction of texts that are classified, since some texts may not contain a single paragraph
for which $|s| \ge T$. To that end, we used the forward test data for 2021 (for the UK) and 
2021-2022 (for Sweden) to compute, for various values of $T$, both the FP rate and the 
fraction of classified text samples, \textit{i.e.}, those for which at least one paragraph had a length 
of at least $k$ (=30) and a classification sum $|s| \ge T$. Note that, for \textit{this} subset,
the ground truth class is~0 (human-written) for all samples, such that any assignment
of class~1 (for a given text) increases the FP rate.
The results are shown in Fig.~\ref{fig:fprates}. As expected, the FP rate falls dramatically 
as $T$ increases. Specifically, using $T = 10$, we can classify around 94\% of the Swedish 
text samples (from the selected subset, \textit{i.e.}, the forward test set for 2021-2022) with an FP rate 
very close to 0. Similarly, for the UK data (for 2021), with the same threshold ($T = 10$) the
FP rate is again close to 0, but in this case only around 82\% of the samples are
classified. However, as can be seen in the left panel of Fig.~\ref{fig:fprates}, for the UK data one can 
reduce the threshold to around 8, while still keeping the FP rate close to 0 and classifying around 
90\% of the samples. Thus, we can conclude that, by slightly sacrificing classifiability, one
can obtain near-zero FP rate also for the paragraph-level classification.

\subsection{Classifier interpretability}
Studies on LLM-generated text detection often highlight features that are more prominent in LLM-generated text. This approach does provide some useful insights and allows one to compare common patterns between human-authored and LLM-generated text. However, such an approach can be misunderstood as an oversimplification of the reasoning behind a classification result: A class label is rarely assigned based on just a few features alone (which is also why text \textit{length} is a very important factor for a reliable classification). Using features that are more common in LLM-generated text does \textit{not} automatically mean that the text is labelled as such, and such claims can also discourage the use of certain words that are perfectly valid to use. Instead, as we will now illustrate, the classification is determined by a larger constellation of features working together.

\begin{figure}
    \centering
    \includegraphics[width=0.8\linewidth]{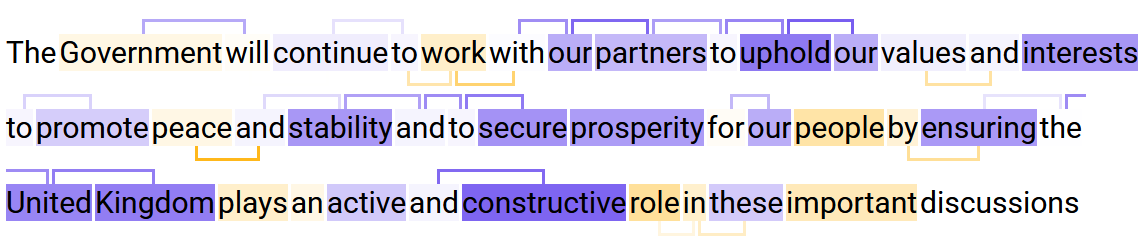}
    \caption{A classified text sample of Class~1 (LLM-generated) from the base test set. The text was labelled correctly as LLM-generated by our classifier. The unigram and bigram features used by the classifier are coloured here in yellow (negative weights) and in purple (positive weights).}
    \label{fig:llmText}
\end{figure}

\begin{figure}
    \centering
    \includegraphics[width=0.8\linewidth]{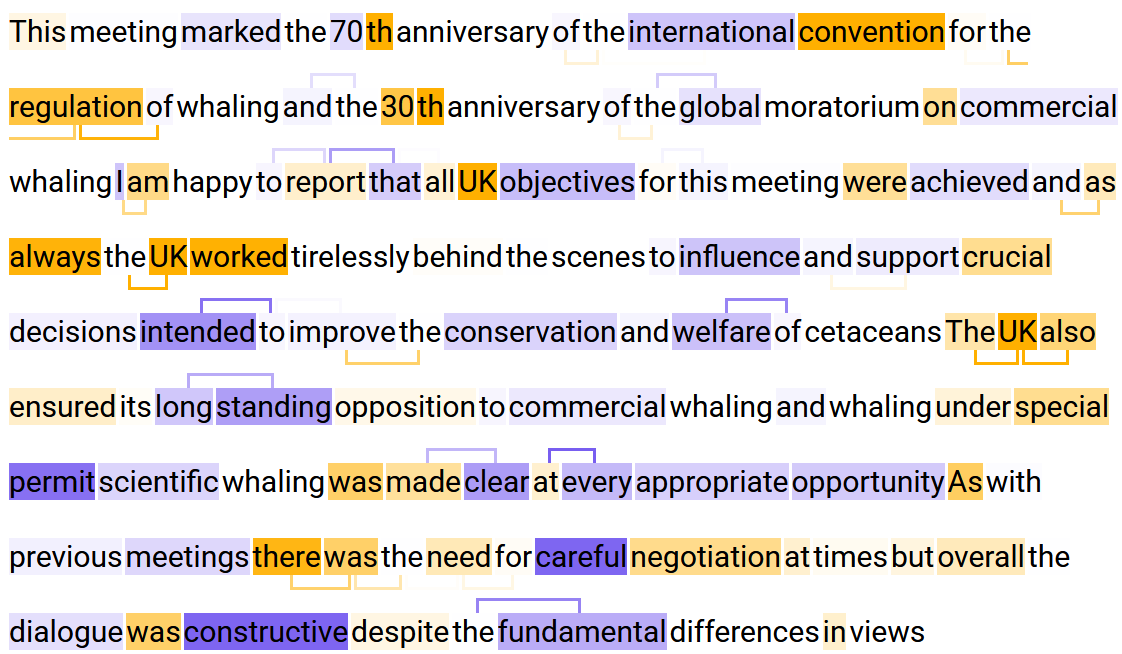}
    \caption{A classified text sample of Class~0 (human-written) from the base test set. The text was labelled correctly as human-written by our classifier. See also the caption for Figure~\ref{fig:llmText}.}
    \label{fig:humanText}
\end{figure}

Visual representations of two classified texts from the base test set are shown in Figures~\ref{fig:llmText} and~\ref{fig:humanText}. In this representation, unigram features (words) are shaded in colour, such that the colours represent the feature weights: Features with negative weights (Class~0, human-written) are coloured in yellow  and features with positive weights (Class~1, LLM-generated) are in purple. The intensity of the colour indicates the magnitude of the feature weight. Bigram features are indicated by arches connecting the two words forming each bigram. Arches indicative of positive weights are shown above the row of text where the bigram appears, whereas arches for negative weights are shown below the text. The colour intensity of an arch is determined by the magnitude of the feature weight in the same way as for the unigrams.

The text in Figure~\ref{fig:llmText} had a classification sum of around 13.6 and was thus correctly classified as LLM-generated.
As can be seen from the figure, the text consists largely of positive features, coloured in purple but, as described above, the positive detection cannot be attributed to a single feature (or just a few) in this example; here, there are also multiple features providing a negative contribution, as well 
as several bigrams, some with positive weights and some with negative weights.
We thus observe that evading the detector would require multiple modifications post-generation; simple synonym replacements would be insufficient for achieving the desired result in this case, and in many other cases.
An example of a correctly classified human-written text with a classification sum of -11.8 is shown in Figure~\ref{fig:humanText}. Similar to the LLM-generated text, the classification is formed by using multiple, in this case predominantly negative, features.
Here, the text also includes some strongly positive features, such as \textit{constructive} (which also appears in the LLM-generated text in Figure~\ref{fig:llmText}). Nevertheless, the combined effect of the features was to correctly classify the text as human-written.

As for individual features, an interesting observation (also visible in the two examples in Figures~\ref{fig:llmText} and~\ref{fig:humanText}) is that the abbreviation \textit{UK} is more common in the human-written texts, whereas the LLM-generated texts prefer \textit{United Kingdom} instead. In 
the LLM-generated text in Figure~\ref{fig:llmText}, the term \textit{United Kingdom} appears once. If
it is replaced by~\textit{UK}, in an attempt to make the text more human-like, the classification sum falls to 8.8, but is still strongly positive, again indicating the limited impact of simple synonym replacements.
The existential \textit{there} is another example of a more human-like feature.

We also tested direct synonym replacements for some of the features with positive weights in the text in Figure~\ref{fig:llmText} to see whether the detection could be shifted towards a negative prediction. The modified text reads: \textit{The Government will continue to work with our \underline{allies} to \underline{protect} our values and interests to \underline{encourage} peace and stability and to \underline{build} prosperity for our people by \underline{guaranteeing} the \underline{UK} plays an active and \underline{vital} role in these important discussions}. With these changes, the classification sum was 2.94, meaning that it would still be detected as LLM-generated, albeit with lower confidence.
We also tried an online tool\footnote{Quillbot's AI Humanizer: \url{https://quillbot.com/ai-humanizer}} to make this text sample more human-like, but the tool made very few modifications in the text and the output text was still consistently detected as LLM-generated, with a sum between 11.2 and 12.8.
We did, however, find that when prompting Gemini to make the text more human-like, it was able to generate versions of the text that had a slightly negative classification sum of about -0.3, where the modifications involved multiple structural changes - but the text became much shorter. These findings pertain to the example discussed here, and further work is required to evaluate to what extent paraphrasing can be used to evade the classifier in general.

It should be noted that, rather than being an approximation, our visualisation is an exact representation of the factors that underlie a given classification, made possible by the transparent nature of the linear classifier. Thus, for instance, in the (rare) cases of false positives, the visualisation can be used for understanding the cause of the misclassification. Moreover, in very \textit{short} texts, it is entirely possible that the classification could be dominated by just a few features; the visualisation can then help in making an informed decision regarding the class assignment.

\subsection{Limitations of this study}

The results reported here are subject to the limitations of our methodology. For example, while Tables~\ref{tab:forwardResultsFull} and \ref{tab:forwardResultsParas} appear to show greater AI use in the 
UK parliament, we note that the results may not be directly comparable, owing to linguistic differences
between English and Swedish, affecting the detectability of generated text in each language, and
differences in the output of the LLMs arising from their differing capabilities in each language.
Moreover, while two LLMs were used to generate English text, only one was used for Swedish.

Also, all texts were generated using current LLM versions; results may have been slightly different if we had used the versions current in each year of study, as of course no texts were generated using future models. 
Furthermore, we have used a simple binary classification, treating all
texts as either human-written or LLM-generated. 
It is more likely that some texts will inhabit a grey
area: For example, a user may post-edit an LLM-generated text to produce a passage with mixed characteristics. Here we do not address issues of localising nor quantifying LLM use within a given text sample.

Nonetheless, our results prove accurate in identifying where there is some LLM use, 
which we have noted is always undisclosed, with a very low rate of false positives. 

\subsection{Implications}
\label{sec:imp}

Our results suggest a steadily growing undisclosed use of LLMs in generating parliamentary texts; by 2026, over 15\% of UK and over 9\% of Swedish 
motions are identified as containing at least one AI-generated paragraph. 
While the public response to the the use of of LLM-generated content in political discourse remains a subject
of discussion, artificial text may weaken public perception of politicians' authenticity.
Moreover, it has been reported that citizens value \textit{authenticity} in politicians more than competence and even integrity~\citep{valgarsson_good_2021} .

Parliamentary use of AI has become a matter of public interest, with one popular UK statistics podcast asking \textit{\lq\lq\/Is it possible to prove that MPs are using AI to write their speeches?\rq\rq}~\citep{tim_harford_bbc_2025}; 
while their discussion centred on the appearance of a few telltale expressions possibly indicative of AI, we have now shown
that a large constellation of features can reliably identify LLM-generated content appearing in parliamentary texts.

While some uses of AI may be uncontroversial, for example, to check for spelling or grammar
errors in written statements, especially if the author is not a native speaker of the parliament's 
language, LLM-use without disclosure and careful oversight raises concerns, as the generated text may not accurately reflect the intentions of the `author';
and while politicians have long been assisted by human speech-writers, transferring this role to AI 
introduces new risks of social biases and outright falsehoods being included, as outlined in Section \ref{sect:politics}. 
It may be pertinent to note that, in the context of court proceedings, \cite{paar_artificial_2025} has demonstrated how \textit{\lq\lq even limited AI support can risk blurring the lines between human and AI decision-making\rq\rq} and
lead to human oversight being reduced to rubber stamping.

There appears to be little adherence to guidelines requiring disclosure of where and how AI is used in parliamentary work;
moreover, these guidelines themselves are often rather vague or even meaningless. 
For example, in~\citep{ipu3} it is stated that parliaments should \textit{\lq\lq clearly communicate when AI outputs are probabilistic rather than factual\rq\rq}, which is a meaningless requirement in the context of LLMs, given that \textit{all} generative AI output is probabilistic. Nonetheless, the guidelines indicate an expectation that AI use does not absolve the writer
from responsibility: The (human) author is
still fully responsible for the content.

Even where AI-generated content is disclosed and then reviewed and corrected by humans, 
there are legitimate concerns about its effect on parliamentary processes.
For example,~\citep{maynard_ai_2026} identifies 
what they call a `cognitive Trojan horse', noting that LLM-generated text often has \textit{\lq\lq characteristics that bypass the cognitive mechanisms humans evolved and learned to
evaluate incoming information\rq\rq}, suggesting that errors introduced by AI may evade human detection. 

Other studies have highlighted the danger of so-called cognitive offloading, where use of AI may discourage humans from applying
sufficient independent critical thought, with ~\citep{klein_extended_2025} reporting that \textit{\lq\lq the frictionless availability of AI-generated answers enables users to systematically bypass \ldots effortful cognitive processes\rq\rq}. A study by~\citep{kosmyna_your_2025} using
electroencephalography (EEG) to record participants' brain activity reported 
measurable cognitive effects of LLM use, noting that \textit{\lq\lq participants showed weaker neural connectivity and under-engagement of alpha and beta networks\rq\rq} 
when they had been accustomed to LLM use in essay-writing tasks; however, the authors do caution against over-interpreting their results.

Drawing these threads together, we recommend that clearer, actionable guidelines, and actual adherence to such guidelines should be a requirement for the responsible use of AI in parliamentary processes.

\section{Conclusion and future work}
\label{sect:conclusion}
In this paper, we studied the undisclosed use of LLMs in parliamentary texts in the UK and in Sweden. We trained an interpretable classifier that obtained a high performance on a held-out test set. A low false positive rate over data predating the era of LLMs shows that the classifier is robust and can reliably detect the use of LLMs in previously unseen texts.

The results over the forward test set, consisting of texts that were published after LLMs were released, show a strong increase in the use of LLMs in parliamentary texts published after 2021, especially for the last 1-2 years. We also noted that, despite strong recommendations to disclose the use of AI, the authors of the texts have consistently failed to disclose where and how LLMs were used. 

The procedure described in this paper can be applied on texts from other parliaments that provide open access data. In future work, we also plan to expand on the approach of evaluating the use of LLMs. In this work, we began doing so by classifying individual paragraphs in texts. One could also inspect parts of paragraphs, such as individual sentences. Since such an approach involves classifying shorter portions of text, it is also essential to further study the reasons for false positive detections.
Exploring rephrasing as an adversarial attack is also an interesting direction for future work.

\section*{Acknowledgement(s)}
This work was partially supported by the Wallenberg AI, Autonomous Systems and Software Program (WASP) funded by the Knut and Alice Wallenberg Foundation.

We acknowledge the UK Parliament (Hansard) and the Swedish Riksdag for making their parliamentary data publicly available under their respective open licences, Open Parliament Licence v3.0, UK (\url{https://www.parliament.uk/site-information/copyright-parliament/open-parliament-licence/}) and Riksdag Open Data, Sweden (\url{https://www.riksdagen.se/sv/dokument-och-lagar/riksdagens-oppna-data/anvandarstod/anvandningsvillkor/}).

\section*{Disclosure statement}

The authors report there are no competing interests to declare. 

As described in Section~\ref{sec:data}, a part of the data was generated using LLMs, as an integral and necessary part of this project aimed at \textit{detecting} LLM use. However, apart from this 
usage as described in the paper, the authors report that generative AI was not otherwise used in their research nor in the preparation of this manuscript. 

\section*{Funding}
The authors declare that no funding was obtained for the reported work.

\section*{Data availability statement}
The data that support the findings of this study are openly available in Zenodo at 
https://doi.org/10.5281/zenodo.20611280.

\bibliographystyle{apacite}
\bibliography{references}

@conference{SuvantoEtal2026,
    author={Minerva Suvanto and Andrea McGlinchey and Mattias Wahde and Peter J. Barclay},
    title={{Interpretable Text Classification Applied to the Detection of {LLM}-Generated Creative Writing}},
    booktitle={{Proceedings of the 18th International Conference on Agents and Artificial Intelligence - Volume 2: ICAART}},
    year={2026},
    pages={1198-1209},
    publisher={SciTePress},
    organization={INSTICC},
    doi={10.5220/0014236000004052},
    isbn={978-989-758-796-2},
    issn={2184-433X},
}

@unpublished{SuvantoWahde2026,
  author    = {Minerva Suvanto and Mattias Wahde},
  title     = {A note on the detection of {LLM}-generated political speeches},
  note      = {Manuscript submitted for publication},
  year      = {2026},
  month     = {January}
}

@article{PickeringEtal2025,
    author = {Pickering, Steven David and Hansen, Martin Ejnar and Sunahara, Yosuke},
    title = {Democracy by algorithm? Public attitudes towards {AI} in parliamentary decision-making in the {UK} and {Japan}},
    journal = {Parliamentary Affairs},
    pages = {gsaf050},
    year = {2025},
    month = {10},
    issn = {0031-2290},
    doi = {10.1093/pa/gsaf050},
    eprint = {https://academic.oup.com/pa/advance-article-pdf/doi/10.1093/pa/gsaf050/64881741/gsaf050.pdf},
}

@article{AdamsEtal2023,
  title={{(Why) is misinformation a problem?}},
  author={Adams, Zo{\"e} and Osman, Magda and Bechlivanidis, Christos and Meder, Bj{\"o}rn},
  journal={Perspectives on Psychological Science},
  volume={18},
  number={6},
  pages={1436--1463},
  year={2023},
  publisher={Sage Publications Sage CA: Los Angeles, CA}
}

@article{BaiEtal2025,
  title={{LLM}-generated messages can persuade humans on policy issues},
  author={Bai, Hui and Voelkel, Jan G and Muldowney, Shane and Eichstaedt, Johannes C and Willer, Robb},
  journal={Nature Communications},
  volume={16},
  number={1},
  pages={6037},
  year={2025},
  publisher={Nature Publishing Group UK London}
}

@article{RettenbergerEtal2025,
  title={Assessing political bias in large language models},
  author={Rettenberger, Luca and Reischl, Markus and Schutera, Mark},
  journal={Journal of Computational Social Science},
  volume={8},
  number={2},
  pages={42},
  year={2025},
  publisher={Springer}
}

@misc{EBU2025,
  title={{News Integrity in {AI} Assistants: An international {PSM} study}},
  year={2025},
  howpublished={\url{https://www.ebu.ch/research/open/report/news-integrity-in-ai-assistants}},
  author={{European Broadcasting Union (EBU)}},
  lastchecked = {22.05.2026}
}

@misc{SOM206,
  title={{Svensk {AI}-opinion 2023 - 2025}},
  year={2026},
  howpublished={\url{https://www.gu.se/sites/default/files/2026-04/Svensk%20AI%20opinion%202023-2025.pdf}},
  author={{SOM institute}},
  number={2026:36},
  lastchecked = {22.05.2026}
}

@misc{IPU1,
title={Guidelines for {AI} in parliaments},
year={2024},
howpublished=
{\url{https://www.ipu.org/resources/publications/reference/2024-12/guidelines-ai-in-parliaments}},
author={{IPU}},
lastchecked={20260526}
}

@misc{IPU2,
title={Guidelines for {AI} in parliaments - Key points for {MPs}},
year={2024},
howpublished=
{\url{https://www.ipu.org/file/21238/download}},
author={{IPU}},
lastchecked={20260526}
}

@misc{IPU3,
title={Ethical principles: Transparency},
year={2024},
howpublished=
{\url{https://www.ipu.org/ai-guidelines/ethical-principles-transparency}},
author={{IPU}},
lastchecked={20260610}
}

@misc{UKParliament1,
title={Generative {AI} framework for {HM} Government},
year={2024},
howpublished={\url{https://assets.publishing.service.gov.uk/media/65c3b5d628a4a00012d2ba5c/6.8558_CO_Generative_AI_Framework_Report_v7_WEB.pdf}},
author={{UK Parliament}},
lastchecked={20260526}
}

@article{Giray2024,
  title={The problem with false positives: AI detection unfairly accuses scholars of AI plagiarism},
  author={Giray, Louie},
  journal={The Serials Librarian},
  volume={85},
  number={5-6},
  pages={181--189},
  year={2024},
  publisher={Taylor \& Francis}
}

@incollection{Gegg2024,
  title={AI detection's high false positive rates and the psychological and material impacts on students},
  author={Gegg-Harrison, Whitney and Quarterman, Claire},
  booktitle={Academic integrity in the age of artificial intelligence},
  pages={199--219},
  year={2024},
  publisher={IGI Global Scientific Publishing}
}

@article{Dalalah2023,
  title={The false positives and false negatives of generative AI detection tools in education and academic research: The case of ChatGPT},
  author={Dalalah, Doraid and Dalalah, Osama MA},
  journal={The International Journal of Management Education},
  volume={21},
  number={2},
  pages={100822},
  year={2023},
  publisher={Elsevier}
}

@article{WuEtal2025,
    title = "A Survey on {LLM}-Generated Text Detection: Necessity, Methods, and Future Directions",
    author = "Wu, Junchao  and
      Yang, Shu  and
      Zhan, Runzhe  and
      Yuan, Yulin  and
      Chao, Lidia Sam  and
      Wong, Derek Fai",
    journal = "Computational Linguistics",
    volume = "51",
    number = "1",
    month = mar,
    year = "2025",
    address = "Cambridge, MA",
    publisher = "MIT Press",
    doi = "10.1162/coli_a_00549",
    pages = "275--338",
}

@inproceedings{PotterEtal2024,
    title = "Hidden Persuaders: {LLM}s' Political Leaning and Their Influence on Voters",
    author = "Potter, Yujin  and
      Lai, Shiyang  and
      Kim, Junsol  and
      Evans, James  and
      Song, Dawn",
    editor = "Al-Onaizan, Yaser  and
      Bansal, Mohit  and
      Chen, Yun-Nung",
    booktitle = "Proceedings of the 2024 Conference on Empirical Methods in Natural Language Processing",
    month = nov,
    year = "2024",
    address = "Miami, Florida, USA",
    publisher = "Association for Computational Linguistics",
    doi = "10.18653/v1/2024.emnlp-main.244",
    pages = "4244--4275",
}

@article{HackenburgEtal2025,
  title={The levers of political persuasion with conversational artificial intelligence},
  author={Hackenburg, Kobi and Tappin, Ben M and Hewitt, Luke and Saunders, Ed and Black, Sid and Lin, Hause and Fist, Catherine and Margetts, Helen and Rand, David G and Summerfield, Christopher},
  journal={Science},
  volume={390},
  number={6777},
  pages={eaea3884},
  year={2025},
  publisher={American Association for the Advancement of Science}
}

@inproceedings{ZamaraevaEtal2025,
    title = "Comparing {LLM}-generated and human-authored news text using formal syntactic theory",
    author = "Zamaraeva, Olga  and
      Flickinger, Dan  and
      Bond, Francis  and
      G{\'o}mez-Rodr{\'i}guez, Carlos",
    editor = "Che, Wanxiang  and
      Nabende, Joyce  and
      Shutova, Ekaterina  and
      Pilehvar, Mohammad Taher",
    booktitle = "Proceedings of the 63rd Annual Meeting of the Association for Computational Linguistics (Volume 1: Long Papers)",
    month = jul,
    year = "2025",
    address = "Vienna, Austria",
    publisher = "Association for Computational Linguistics",
    doi = "10.18653/v1/2025.acl-long.443",
    pages = "9041--9060",
    ISBN = "979-8-89176-251-0",
}

@article{ReinhartEtal2025,
author = {Alex Reinhart  and Ben Markey  and Michael Laudenbach  and Kachatad Pantusen  and Ronald Yurko  and Gordon Weinberg  and David West Brown },
title = {Do {LLM}s write like humans? {Variation} in grammatical and rhetorical styles},
journal = {Proceedings of the National Academy of Sciences},
volume = {122},
number = {8},
pages = {e2422455122},
year = {2025},
doi = {10.1073/pnas.2422455122},
eprint = {https://www.pnas.org/doi/pdf/10.1073/pnas.2422455122},
}

@inproceedings{QuidwaiEtal2023,
    title = "Beyond Black Box {AI} generated Plagiarism Detection: From Sentence to Document Level",
    author = "Quidwai, Ali  and
      Li, Chunhui  and
      Dube, Parijat",
    editor = {Kochmar, Ekaterina  and
      Burstein, Jill  and
      Horbach, Andrea  and
      Laarmann-Quante, Ronja  and
      Madnani, Nitin  and
      Tack, Ana{\"i}s  and
      Yaneva, Victoria  and
      Yuan, Zheng  and
      Zesch, Torsten},
    booktitle = "Proceedings of the 18th Workshop on Innovative Use of NLP for Building Educational Applications (BEA 2023)",
    month = jul,
    year = "2023",
    address = "Toronto, Canada",
    publisher = "Association for Computational Linguistics",
    doi = "10.18653/v1/2023.bea-1.58",
    pages = "727--735",
}

@inproceedings{WeiEtal2025,
    title = "Learning to Rewrite: Generalized {LLM}-Generated Text Detection",
    author = "Hao, Wei  and
      Li, Ran  and
      Zhao, Weiliang  and
      Yang, Junfeng  and
      Mao, Chengzhi",
    editor = "Che, Wanxiang  and
      Nabende, Joyce  and
      Shutova, Ekaterina  and
      Pilehvar, Mohammad Taher",
    booktitle = "Proceedings of the 63rd Annual Meeting of the Association for Computational Linguistics (Volume 1: Long Papers)",
    month = jul,
    year = "2025",
    address = "Vienna, Austria",
    publisher = "Association for Computational Linguistics",
    doi = "10.18653/v1/2025.acl-long.322",
    pages = "6421--6434",
    ISBN = "979-8-89176-251-0",
}

@article{WahdeEtal2024,
  title={An interpretable method for automated classification of spoken transcripts and written text},
  author={Wahde, Mattias and Della Vedova, Marco L and Virgolin, Marco and Suvanto, Minerva},
  journal={Evolutionary Intelligence},
  volume={17},
  number={1},
  pages={609--621},
  year={2024},
  publisher={Springer}
}

@incollection{DaSilvaGameiro2024,
  title={LLM detectors},
  author={Da Silva Gameiro, Henrique},
  booktitle={Large Language Models in Cybersecurity: Threats, Exposure and Mitigation},
  pages={197--204},
  year={2024},
  publisher={Springer}
}

@inproceedings{KoniarisEtAl2025,
  title={{ParliaBench}: An Evaluation and Benchmarking Framework for {LLM}-Generated Parliamentary Speech},
  author={Koniaris, Marios and Tsipi, Argyro and Tsanakas, Panayiotis},
  booktitle = {{Proceedings of the Fifteenth Language Resources and Evaluation Conference (LREC 2026)}},
  pages = {4797–4818},
  publisher ={European Language Resources Association (ELRA)},
  doi={https://doi.org/10.63317/447dqkef7ks7},
  year={2026}
}

@article{VonDerHeyde2025,
  title={{Vox Populi, Vox AI? Using Language Models to Estimate German Public Opinion}},
  author={Von Der Heyde, Leah and Haensch, Anna-Carolina and Wenz, Alexander},
  journal={Social Science Computer Review},
  pages={08944393251337014},
  year={2025},
  publisher={SAGE Publications Sage CA: Los Angeles, CA}
}

@article{Rozado2024,
  title={The political preferences of LLMs},
  author={Rozado, David},
  journal={PloS one},
  volume={19},
  number={7},
  pages={e0306621},
  year={2024},
  publisher={Public Library of Science}
}

@inproceedings{Macko2023,
  title={MULTITuDE: Large-scale multilingual machine-generated text detection benchmark},
  author={Macko, Dominik and Moro, Robert and Uchendu, Adaku and Lucas, Jason and Yamashita, Michiharu and Pikuliak, Mat{\'u}{\v{s}} and Srba, Ivan and Le, Thai and Lee, Dongwon and Simko, Jakub and others},
  booktitle={Proceedings of the 2023 Conference on Empirical Methods in Natural Language Processing},
  pages={9960--9987},
  year={2023}
}

@inproceedings{Wang2024,
  title={M4: Multi-generator, multi-domain, and multi-lingual black-box machine-generated text detection},
  author={Wang, Yuxia and Mansurov, Jonibek and Ivanov, Petar and Su, Jinyan and Shelmanov, Artem and Tsvigun, Akim and Whitehouse, Chenxi and Afzal, Osama Mohammed and Mahmoud, Tarek and Sasaki, Toru and others},
  booktitle={Proceedings of the 18th Conference of the European Chapter of the Association for Computational Linguistics (Volume 1: Long Papers)},
  pages={1369--1407},
  year={2024}
}

@inproceedings{Salem2026,
  title={AI-Generated Text Detector for Arabic Language: Survey and Remarks},
  author={Salem, Hamza and Elkhateb, Abdelkareem Gaballah and Zafar, Muhammad Naveed and Mazzara, Manuel and Al-Atyat, Yazan Khalaf and Hattab, Siham},
  booktitle={International Conference on Advanced Information Networking and Applications},
  pages={171--181},
  year={2026},
  organization={Springer}
}

@inproceedings{Mao2025,
  title={Mlsdet: Multi-llm statistical deep ensemble for chinese ai-generated text detection},
  author={Mao, Dianhui and Zhang, Denghui and Zhang, Ao and Zhao, Zhihua},
  booktitle={ICASSP 2025-2025 IEEE International Conference on Acoustics, Speech and Signal Processing (ICASSP)},
  pages={1--5},
  year={2025},
  organization={IEEE}
}

@inproceedings{McGlinchey2025,
	address = {Porto, Portugal},
    booktitle={Proceedings of the 17th International Conference on Agents and Artificial Intelligence (ICAART)},
	title = {Using {Machine} {Learning} to {Distinguish} {Human}-{Written} from {Machine}-{Generated} {Creative} {Fiction}.},
	volume = {2},
	isbn = {978-989-758-737-5},
	language = {English},
	author = {Andrea Cristina McGlinchey and Peter J. Barclay},
	month = Feb,
	year = {2025},
	pages = {79--90}
}

@inproceedings{Clark2021,
  title={All that’s ‘human’ is not gold: Evaluating human evaluation of generated text},
  author={Clark, Elizabeth and August, Tal and Serrano, Sofia and Haduong, Nikita and Gururangan, Suchin and Smith, Noah A},
  booktitle={Proceedings of the 59th Annual Meeting of the Association for Computational Linguistics and the 11th International Joint Conference on Natural Language Processing (Volume 1: Long Papers)},
  pages={7282--7296},
  year={2021}
}

@article{Richter2026,
  title={Can humans actually be trusted to detect {LLM}-generated texts? {Assessing} reviewers’ accuracy and confidence},
  author={Richter, Michelle and Larsson, Tove},
  journal={Research Methods in Applied Linguistics},
  volume={5},
  number={2},
  pages={100322},
  year={2026},
  publisher={Elsevier}
}

@article{Munoz2024,
  title={Contrasting linguistic patterns in human and LLM-generated news text},
  author={Mu{\~n}oz-Ortiz, Alberto and G{\'o}mez-Rodr{\'\i}guez, Carlos and Vilares, David},
  journal={Artificial Intelligence Review},
  volume={57},
  number={10},
  pages={265},
  year={2024},
  publisher={Springer}
}

@inproceedings{Zeleke2025,
  title={Human or GenAI? Characterizing the Linguistic Differences between Human-Written and LLM-Generated Text},
  author={Zeleke, Brook and Soni, Amish and Manikonda, Lydia},
  booktitle={Companion Publication of the 17th ACM Web Science Conference 2025},
  pages={34--37},
  year={2025}
}

@article{Herbold2023,
  title={A large-scale comparison of human-written versus ChatGPT-generated essays},
  author={Herbold, Steffen and Hautli-Janisz, Annette and Heuer, Ute and Kikteva, Zlata and Trautsch, Alexander},
  journal={Scientific reports},
  volume={13},
  number={1},
  pages={18617},
  year={2023},
  publisher={Nature Publishing Group UK London}
}

@inproceedings{TaubenfeldEtal2024,
    title = "Systematic Biases in {LLM} Simulations of Debates",
    author = "Taubenfeld, Amir  and
      Dover, Yaniv  and
      Reichart, Roi  and
      Goldstein, Ariel",
    editor = "Al-Onaizan, Yaser  and
      Bansal, Mohit  and
      Chen, Yun-Nung",
    booktitle = "Proceedings of the 2024 Conference on Empirical Methods in Natural Language Processing",
    month = nov,
    year = "2024",
    address = "Miami, Florida, USA",
    publisher = "Association for Computational Linguistics",
    doi = "10.18653/v1/2024.emnlp-main.16",
    pages = "251--267",
}

@inproceedings{WanEtal2023,
    title = "{``Kelly is a Warm Person, Joseph is a Role Model'': Gender Biases in {LLM}-Generated Reference Letters}",
    author = "Wan, Yixin  and
      Pu, George  and
      Sun, Jiao  and
      Garimella, Aparna  and
      Chang, Kai-Wei  and
      Peng, Nanyun",
    editor = "Bouamor, Houda  and
      Pino, Juan  and
      Bali, Kalika",
    booktitle = "Findings of the Association for Computational Linguistics: EMNLP 2023",
    month = dec,
    year = "2023",
    address = "Singapore",
    publisher = "Association for Computational Linguistics",
    doi = "10.18653/v1/2023.findings-emnlp.243",
    pages = "3730--3748",
}

@inproceedings{VermaEtal2024,
    title = "Ghostbuster: Detecting Text Ghostwritten by Large Language Models",
    author = "Verma, Vivek  and
      Fleisig, Eve  and
      Tomlin, Nicholas  and
      Klein, Dan",
    editor = "Duh, Kevin  and
      Gomez, Helena  and
      Bethard, Steven",
    booktitle = "Proceedings of the 2024 Conference of the North American Chapter of the Association for Computational Linguistics: Human Language Technologies (Volume 1: Long Papers)",
    month = jun,
    year = "2024",
    address = "Mexico City, Mexico",
    publisher = "Association for Computational Linguistics",
    doi = "10.18653/v1/2024.naacl-long.95",
    pages = "1702--1717",
}

@article{WuEtal2024,
  title={Detectrl: Benchmarking llm-generated text detection in real-world scenarios},
  author={Wu, Junchao and Zhan, Runzhe and Wong, Derek F and Yang, Shu and Yang, Xinyi and Yuan, Yulin and Chao, Lidia S},
  journal={Advances in Neural Information Processing Systems},
  volume={37},
  pages={100369--100401},
  year={2024}
}

@article{ChengEtal2026,
  title={Adversarial Paraphrasing: A Universal Attack for Humanizing AI-Generated Text},
  author={Cheng, Yize and Sadasivan, Vinu Sankar and Saberi, Mehrdad and Saha, Shoumik and Feizi, Soheil},
  journal={Advances in Neural Information Processing Systems},
  volume={38},
  pages={47591--47622},
  year={2026}
}

@article{Min2025,
 author  = {Min, Roselyne},
 year={2025},
 date    = {2025-08-07},
 title   = {{ Sweden’s prime minister uses ChatGPT. How else are governments using chatbots?}},
 journal = {Euronews},
 url     = {https://www.euronews.com/next/2025/08/07/swedens-prime-minister-uses-chatgpt-how-else-are-governments-using-chatbots},
 lastchecked = {08.06.2026}
}

@article{Rudin2019,
  title={Stop explaining black box machine learning models for high stakes decisions and use interpretable models instead},
  author={Rudin, Cynthia},
  journal={Nature machine intelligence},
  volume={1},
  number={5},
  pages={206--215},
  year={2019},
  publisher={Nature Publishing Group UK London}
}

@article{klein_extended_2025,
	title = {The extended hollowed mind: why foundational knowledge is indispensable in the age of {AI}},
	volume = {8},
	shorttitle = {The extended hollowed mind},
	urldate = {2026-06-09},
	journal = {Frontiers in Artificial Intelligence},
	publisher = {Frontiers Media SA},
	author = {Klein, Christian R. and Klein, Reinhard},
	year = {2025},
	pages = {1719019},
}

@misc{maynard_ai_2026,
	title = {The {AI} {Cognitive} {Trojan} {Horse}: {How} {Large} {Language} {Models} {May} {Bypass} {Human} {Epistemic} {Vigilance}},
	shorttitle = {The {AI} {Cognitive} {Trojan} {Horse}},
	doi = {10.48550/arXiv.2601.07085},
	urldate = {2026-06-09},
	publisher = {arXiv},
	author = {Maynard, Andrew D.},
	month = may,
	year = {2026},
	note = {arXiv:2601.07085 [cs.HC]},
}

@inproceedings{mansour_sami_case_2023,
	address = {Bogotá, Colombia},
	series = {2576-3148},
	title = {A case study of {Fairness} in generated images of {Large} {Language} {Models} for {Software} {Engineering} {Tasks}.},
	isbn = {979-8-3503-2783-0},
	url = {https://ieeexplore.ieee.org/stamp/stamp.jsp?tp=&arnumber=10336315&tag=1},
	doi = {https://doi.org/10.1109/icsme58846.2023.00051},
	booktitle = {Proceedings of the 39th {IEEE} {International} {Conference} on {Software} {Maintenance} and {Evolution} ({ICSME} 2023).},
	publisher = {IEEE},
	author = {Mansour Sami and Ashkan Sami and Peter J. Barclay},
	month = Oct,
	year = {2023},
	pages = {391--396},
}

@inproceedings{barclay_investigating_2024,
	address = {Rovaniemi, Finland},
	title = {Investigating {Markers} and {Drivers} of {Gender} {Bias} in {Machine} {Translations}},
	doi = {10.1109/SANER60148.2024.00054},
	language = {English},
	booktitle = {Proceedings of {SANER24}},
	publisher = {IEEE},
	author = {Barclay, Peter and Sami, Ashkan},
	month = mar,
	year = {2024},
	pages = {455 -- 464},
}

@article{valgarsson_good_2021,
	title = {The {Good} {Politician} and {Political} {Trust}: {An} {Authenticity} {Gap} in {British} {Politics}?},
	volume = {69},
	issn = {0032-3217, 1467-9248},
	shorttitle = {The {Good} {Politician} and {Political} {Trust}},
	doi = {10.1177/0032321720928257},
	language = {en},
	number = {4},
	urldate = {2026-05-27},
	journal = {Political Studies},
	author = {Valgarðsson, Viktor Orri and Clarke, Nick and Jennings, Will and Stoker, Gerry},
	month = nov,
	year = {2021},
	pages = {858--880},
}

@misc{tim_harford_bbc_2025,
	type = {podcast},
	title = {{BBC} {Radio} 4 - {More} or {Less}.},
	url = {https://www.bbc.co.uk/programmes/m002jsw9},
	language = {en-GB},
	urldate = {2026-05-28},
	journal = {BBC},
	publisher = {BBC},
	author = {{BBC Sound}},
	month = sep,
	year = {2025},
	note = {Available online from the {UK} only. Presenter: Tim Harford
Reporter: Lizzy McNeill
Producers: Nathan Gower and Nicholas Barrett
Series producer: Tom Colls
Production co-ordinator: Maria Ogundele
Sound mix: Gareth Jones
Editor: Richard Vadon},
}

@article{kosmyna_your_2025,
	title = {Your brain on {ChatGPT}: {Accumulation} of cognitive debt when using an {AI} assistant for essay writing task},
	volume = {4},
	shorttitle = {Your brain on {ChatGPT}},
	urldate = {2026-06-09},
	journal = {arXiv preprint arXiv:2506.08872},
	author = {Kosmyna, Nataliya and Hauptmann, Eugene and Yuan, Ye Tong and Situ, Jessica and Liao, Xian-Hao and Beresnitzky, Ashly Vivian and Braunstein, Iris and Maes, Pattie},
	year = {2025},
}

@article{paar_artificial_2025,
	title = {Artificial {Intelligence} in {Court} {Proceedings}: {Judge}’s {Little} {Helper} or the {Beginning} of {AI}’s {Hostile} {Takeover}?},
	volume = {26},
	shorttitle = {Artificial {Intelligence} in {Court} {Proceedings}},
	url = {https://www.cambridge.org/core/journals/german-law-journal/article/artificial-intelligence-in-court-proceedings-judges-little-helper-or-the-beginning-of-ais-hostile-takeover/EB07B24BD131360172F24086B659FA71},
	number = {7},
	urldate = {2026-06-11},
	journal = {German Law Journal},
	publisher = {Cambridge University Press},
	author = {Paar, Elisabeth},
	year = {2025},
	pages = {1383--1403},
}

\section{Appendices}

\appendix
\section{Prompts}
\label{appendix:prompts}

The prompts that were sent to the LLMs in order to generate texts for the base data set are listed here. The following placeholders were replaced by string values in the final prompts: \verb+MAX_SENTS+, \verb+MAX_TOKENS+, \verb+FULL_TXT+, and \verb+SUMMARY_TXT+

\subsection{Swedish text summarisation}

\begin{verbatim}
Du är en expert på att sammanfatta texter. Din uppgift är att skapa en 
högkvalitativ sammanfattning som fångar huvudtesen, stödjande argument
och kritiska nyanser. Svara enbart med sammanfattningen i form av en 
punktlista med maximalt MAX_SENTS punkter.
Använd ett enkelt och okomplicerat språk samt en neutral ton. Undvik att
skriva i tredje person. Håll hela sammanfattningen under MAX_TOKENS ord 
totalt. Skriv fullständiga meningar utan några specialtecken.

Text: FULL_TXT

Sammanfattning:
\end{verbatim}

\subsection{English text summarisation}

\begin{verbatim}
You are an Executive text summarizer. Your goal is to provide a 
high-fidelity summary of the provided text that captures the core 
thesis, supporting arguments, and critical nuances. Respond only with 
the summary in the form of bullet points with a maximum of MAX_SENTS points.
Write in standard British English language and use simple, non-complex 
language. Avoid using the third person. Keep the whole summary below 
MAX_TOKENS words. Write full sentences without any special characters.

Text: FULL_TXT

Summary:
\end{verbatim}

\subsection{Swedish full text generation}

\begin{verbatim}
Du är ledamot av Sveriges riksdag. Skriv en riksdagsmotion baserat på 
nedanstående sammanfattning. Använd ett korrekt, formellt och parlamentariskt 
språkbruk. Dela upp texten i stycken, men utan rubriker för styckena. Ange 
inte datum eller författarnamn. Svara endast med motionen, inga förklaringar, 
kommentarer eller anmärkningar, ej heller förslag på fortsatta diskussioner.

Sammanfattning: SUMMARY_TXT

\end{verbatim}

\subsection{English full text generation}

\begin{verbatim}
You are a UK parliamentary member. Generate a coherent political statement 
based on the provided summary. Use proper British parliamentary language. 
Divide the text into paragraphs without headings for the paragraphs. Do not 
include dates or author names. Respond only with the statement, no explanations, 
comments or remarks, or suggestions for further discussions.

Summary: SUMMARY_TXT
\end{verbatim}

\end{document}